\def\year{2022}\relax
\documentclass[letterpaper]{article} % DO NOT CHANGE THIS
\usepackage{aaai22}  % DO NOT CHANGE THIS
\usepackage{times}  % DO NOT CHANGE THIS
\usepackage{helvet}  % DO NOT CHANGE THIS
\usepackage{courier}  % DO NOT CHANGE THIS
\usepackage[hyphens]{url}  % DO NOT CHANGE THIS
\usepackage{graphicx} % DO NOT CHANGE THIS
\usepackage{natbib}  % DO NOT CHANGE THIS AND DO NOT ADD ANY OPTIONS TO IT
\usepackage{caption} % DO NOT CHANGE THIS AND DO NOT ADD ANY OPTIONS TO IT
\usepackage{cite}
\usepackage{amsmath}
\DeclareCaptionStyle{ruled}{labelfont=normalfont,labelsep=colon,strut=off} % DO NOT CHANGE THIS
\usepackage{algorithm}
\usepackage{algorithmic}
\usepackage{subfigure}
\usepackage{threeparttable}
\usepackage{multirow}
\usepackage{booktabs}
\usepackage{xcolor}

\usepackage{newfloat}
\usepackage{listings}
\usepackage[switch]{lineno}

\floatstyle{ruled}
\newfloat{listing}{tb}{lst}{}
\floatname{listing}{Listing}
\title{Activation Modulation and Recalibration Scheme \\ for Weakly Supervised Semantic Segmentation}

\author{
    Jie Qin \textsuperscript{\rm 1,2,3} \footnote{This work was done when Jie Qin interned at Bytedance Inc.},
    Jie Wu \textsuperscript{\rm 2},
    Xuefeng Xiao \textsuperscript{\rm 2},
    Lujun Li \textsuperscript{\rm 3},
    Xingang Wang \textsuperscript{\rm 3}
}
\affiliations{
    \textsuperscript{\rm 1} School of Artificial Intelligence, University of Chinese Academy of Sciences \\
    \textsuperscript{\rm 2} ByteDance Inc
    \textsuperscript{\rm 3} Institute of Automation, Chinese Academy of Sciences\\
    \{qinjie2019, lilujun2019,
 xingang.wang\}@ia.ac.cn, \{wujie.10,
xiaoxuefeng.ailab\}@bytedance.com
}

\usepackage{pifont}

\newcommand{\cmark}{\ding{51}}%

\begin{document}

\makeatletter
\DeclareRobustCommand\onedot{\futurelet\@let@token\@onedot}
\def\@onedot{\ifx\@let@token.\else.\null\fi\xspace}

\def\eg{\emph{e.g}\onedot} \def\Eg{\emph{E.g}\onedot}
\def\ie{\emph{i.e}\onedot} \def\Ie{\emph{I.e}\onedot}
\def\cf{\emph{c.f}\onedot} \def\Cf{\emph{C.f}\onedot}
\def\etc{\emph{etc}\onedot} \def\vs{\emph{vs}\onedot}
\def\wrt{w.r.t\onedot} \def\dof{d.o.f\onedot}
\def\etal{\emph{et al}\onedot}
\makeatother

\maketitle

\begin{abstract}
    Image-level weakly supervised semantic segmentation (WSSS) is a fundamental yet challenging computer vision task facilitating scene understanding and automatic driving.
    Most existing methods resort to classification-based Class Activation Maps (CAMs) to play as the initial pseudo labels, which tend to focus on the discriminative image regions and lack customized characteristics for the segmentation task.
    To alleviate this issue, we propose a novel activation modulation and recalibration (AMR) scheme, which leverages a spotlight branch and a compensation branch to obtain weighted CAMs that can provide recalibration supervision and task-specific concepts.
    Specifically, an attention modulation module (AMM) is employed to rearrange the distribution of feature importance from the channel-spatial sequential perspective, which helps to explicitly model channel-wise interdependencies and spatial encodings to adaptively modulate segmentation-oriented activation responses.
    Furthermore, we introduce a cross pseudo supervision for dual branches, which can be regarded as a semantic similar regularization to mutually refine two branches. 
    Extensive experiments show that AMR establishes a new state-of-the-art performance on the PASCAL VOC 2012 dataset, surpassing not only current methods trained with the image-level of supervision but also some methods relying on stronger supervision, such as saliency label. Experiments also reveal that our scheme is plug-and-play and can be incorporated with other approaches to boost their performance. Our code is available at: \url{https://github.com/jieqin-ai/AMR}
\end{abstract}

\section{Introduction}
Semantic segmentation is a fundamental and crucial task due to extensive applications in the field of computer vision. It aims to perform a pixel-level prediction to cluster parts of an image together that belong to the same object class. 
Albeit with varying degrees of progress, most of its recent successes ~\cite{chen2017deeplab, chen2018encoder} are involved in a fully supervised setting. It is still arduous to acquire such granular pixel-level annotations that require a huge amount of manual effort. To alleviate such expensive and unwieldy annotations, many works tend to resort to weakly supervised manner \cite{wu2020reinforcement,wu2021weakly}, such as bounding boxes supervision~\cite{dai2015boxsup}, scribbles supervision~\cite{lin2016scribblesup}, points supervision~\cite{bearman2016s}, and image-level supervision~\cite{chang2020weakly, ahn2018learning}. Image-level weak supervision is an exceedingly favorable scheme since such coarse annotations are consistent with reality as such weak labels are more readily available in practice. 
In our work, we focus on the image-level weakly supervised paradigm.

\begin{figure}[t]
    \centering
    \begin{center}
        \includegraphics[width=1.0\linewidth]{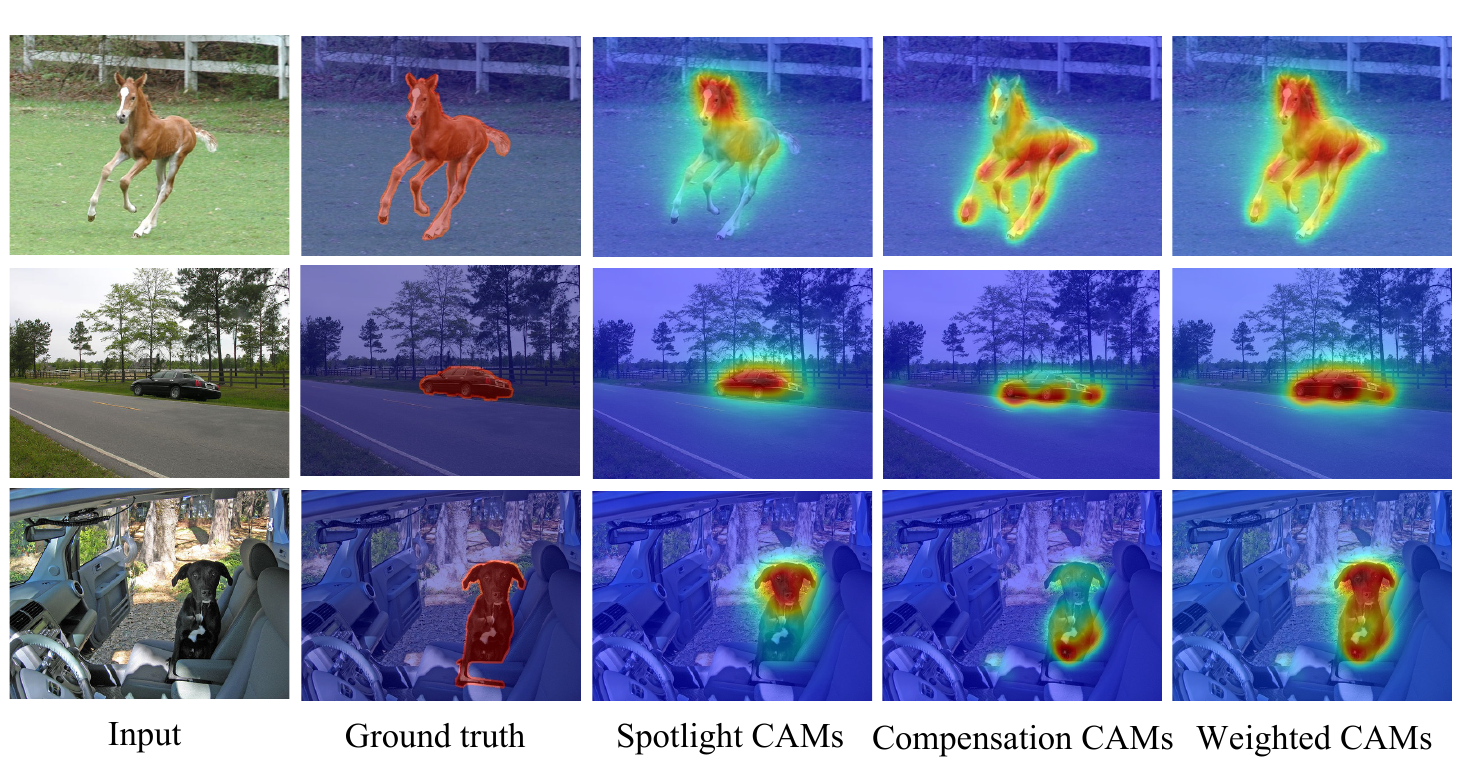}
    \end{center}
    \caption{Visualizations of CAMs in the AMR scheme. ``Spotlight CAMs'' focus more on the discriminative regions similar to conventional CAMs. ``Compensation CAMs'' help to dig out the important but easily ignored regions. The spotlight CAMs are recalibrated by the compensation CAMs and further obtain the ``Weighted CAMs'', which contribute to providing more segmentation-oriented concepts.}
    \label{fig: cam_generate}
  \end{figure}
 Previous image-level WSSS works~\cite{lee2019ficklenet, singh2017hide, wang2020self, choe2020attention} mostly employ classification networks to generate the Class Activation Maps (CAMs)~\cite{zhou2016learning} as the initial pseudo labels for segmentation.  However, this kind of CAM is oriented for classification and lacks customized optimization for the segmentation characteristics. Namely, the classifiers appear to highlight the most discriminative regions, hence the obtained CAM seeds only cover part of the target objects that are consistent with the spotlight CAMs in Fig.~\ref{fig: cam_generate}.
To address this issue, some approaches attempt to expand the discriminative response regions and refine the initial CAM seeds.
SEAM~\cite{wang2020self} adds equivariance regularization on different transformed images to acquire more seed regions. Similarly, ~\cite{wei2017object} presses the model to concentrate on the other regions by iteratively erasing the seeds of CAMs. 
However, these methods usually formulate the expanding process as a complex training stage, \emph{e.g.} the iterative erasing manner is time-consuming and difficult to determine the best number of iterations. Furthermore, it heavily relies on the discriminative regions provided via the classification networks, which easily fails to take the minor important regions into account.
    
To better cope with the above issues, we propose a novel \textbf{A}ctivation \textbf{M}odulation and \textbf{R}ecalibration scheme, termed AMR. The scheme leverages a \emph{spotlight branch} and a \emph{compensation branch} to provide complementary and task-oriented CAMs for WSSS. The spotlight branch denotes the fundamental classification network to produce CAMs, which usually highlight the discriminative and classification-specific regions, such as the head of horse and the window of the car (refer to Fig.~\ref{fig: cam_generate}).
AMR alleviates the task gap issue of using classification-based CAMs to perform segmentation tasks in previous works, which contributes to providing more semantic segmentation-specific cues. Moreover, an attention modulation module (AMM) is employed to rearrange the distribution of activation importance from the channel-spatial sequential perspective, which contributes to modulating segmentation-oriented activation responses adaptively.
 The contributions of AMR can be summarized as follows:    
    
 \begin{itemize}
 \item To the best of our knowledge, we offer the first attempt to explore a plug-and-play compensation branch to provide complementary supervision and task-specific CAMs in WSSS. The compensation branch can dig out the essential regions for segmentation  (such as the legs of the horse and the chassis of the car in Fig.~\ref{fig: cam_generate}), which is very critical to break through the bottleneck of classification-based CAMs for applying in the segmentation task. The compensation CAMs assist in generating the segmentation-oriented CAMs by recalibrating the spotlight CAMs.  Additionally, we introduce a cross pseudo supervision to optimize the output CAMs from dual branches, which can be viewed as the semantic similar regularization to avoid the compensation CAMs concentrating on the background and force it close to spotlight CAMs. 
 
 \item We design an attention modulation module (AMM), which encourages the activation maps to pay equal attention to the whole target objects by performing feature modulation in the channel and spatial dimensions sequentially. A modulation function is leveraged to rearrange the distribution of activation features, which attempts to emphasize minor features and penalize the saliency features that have been captured by the spotlight branch. 
 The channel-spatial sequential manner contributes to explicitly modelling channel-wise interdependencies and spatial encodings within local receptive fields at each layer to adaptively modulate segmentation-oriented features responses.
 
 \item Our approach achieves 68.8\% and 69.1\% in terms of mIoU on validation and test set, which establishes a new state-of-the-art performance in WSSS on the PASCAL VOC2012 dataset~\cite{everingham2015pascal}. Extensive experiments show that AMR surpasses not only current methods trained with the image-level supervision but also some methods relying on stronger supervision, such as saliency label. Experiments also reveal that our scheme is plug-and-play and can can be incorporated with other approaches to boost their performance.
 \end{itemize}

\begin{figure*}[t!]
    \centering
    \begin{center}
        \includegraphics[width=1.0\textwidth]{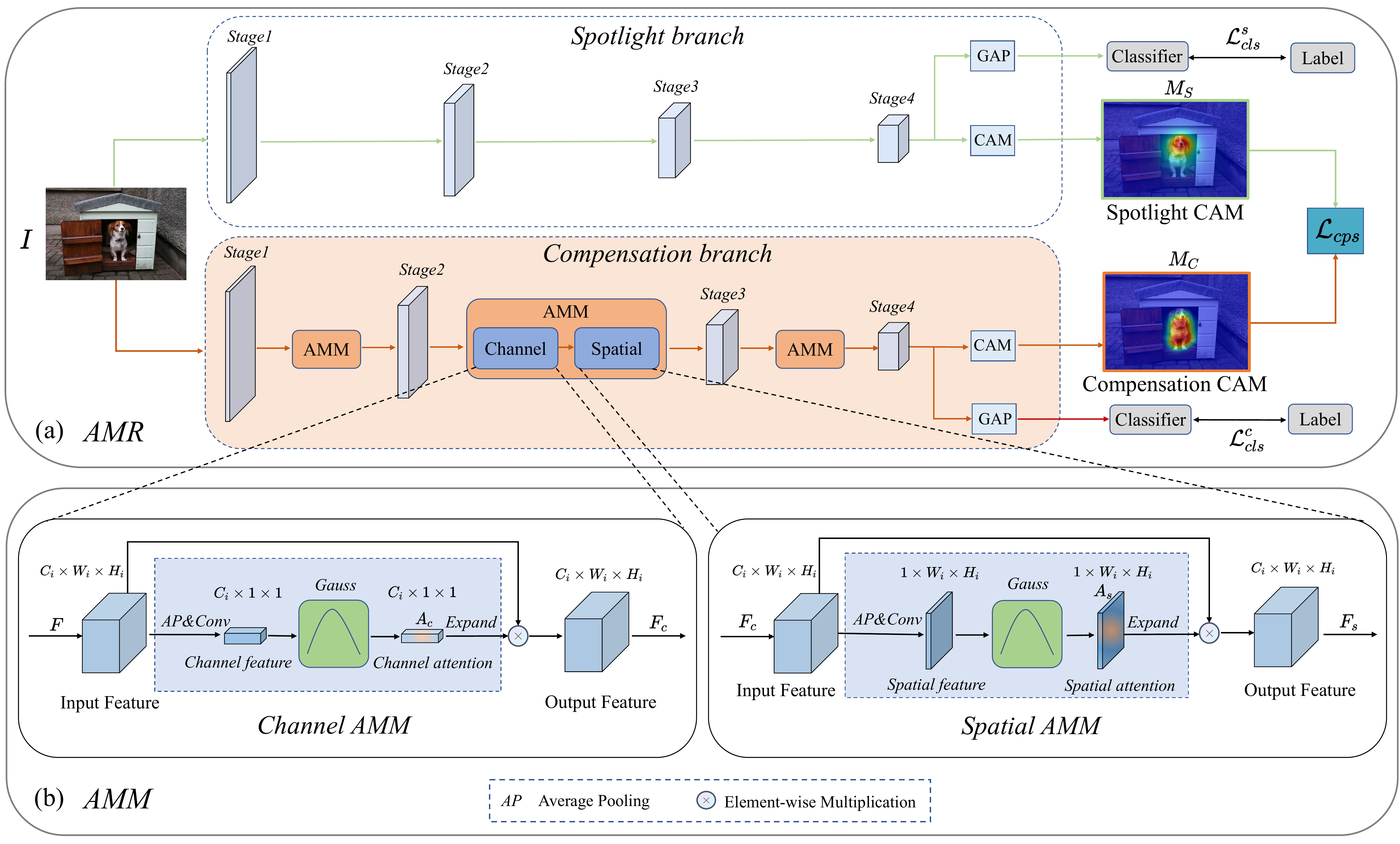}
    \end{center}
    \caption{The framework of the AMR scheme. (a) represents the whole pipeline of AMR. AMR consists of dual branches, \emph{i.e}. \emph{spotlight branch} and \emph{compensation branch}. ``GAP'' represents the global average pooling. (b) Illustration of the AMM, which aims to modulate the activation maps of features in the channel-spatial sequential manner. }
    \label{fig: framework}
  \end{figure*}
 
\section{Related Work}
\subsection{Weakly Supervised Semantic Segmentation}

With the refined research of semantic segmentation, on the one hand, AutoML \cite{li2021revisiting, ren2021online,li2020pams,li2019oicsr,xiao2017design,xia2020hnas} based technologies are employed to improve the segmentation quality. On the other hand, training with lightweight annotation cost is widely explored, image-level WSSS has been extensively studied in recent years. 
Existing advanced methods usually rely on the seed area of Class Activation Maps(CAMs)~\cite{zhou2016learning} generated by the classification networks. Most of these efforts~\cite{wei2018revisiting, wang2020self, sun2020mining, lee2021anti, ahn2018learning, jiang2019integral, wu2021embedded, liu2020leveraging, oh2021background, shen2021parallel} can be classified in two aspects: \emph{generating high-quality CAM seeds} and \emph{refining the pseudo labels}. 
On the one hand, some approaches directly expand the response regions of CAMs because the original activation maps only highlight the discriminative regions of the images. ~\cite{wei2018revisiting} uses dilated convolution with different dilate rates to increase the target regions.
OAA~\cite{jiang2019integral} fuses multi-attention maps in different training processes.
SEAM~\cite{wang2020self} captures different regions from transformed images via equivariance regularization in classification networks.
~\cite{chang2020weakly} explores the feature learning ability of the sub-categories of annotated classes.
~\cite{fan2020cian} and ~\cite{sun2020mining} capture the information of cross-image semantic similarities and differences.
On the other hand, some works focus on refining the pseudo labels based on the initial CAMs. SEC~\cite{kolesnikov2016seed} explores three principles to refine the seeds, i.e., seed, expansion, and constraining.
% DSRG~\cite{huang2018weakly} expands the seed regions by using seed region growing. 
AffinityNet~\cite{ahn2018learning} learns the relation of pixels and propagates the similar semantic pixels by a random walk algorithm. 
In addition, several methods~\cite{yao2021non, lee2019ficklenet} take the CAMs as foreground cues and saliency maps \cite{zhang2019training} as background cues. ~\cite{yao2021non} introduces a graph-based global reasoning unit to discover the objects in the non-salient regions. 
% ~\cite{wei2017object} enforces CNN to focus on more regions by erasing the most discriminative regions continuously.
FickleNet~\cite{lee2019ficklenet} randomly selects the hidden units in the feature maps to discover the other part of objects. 
However, these approaches are formulated in an iterative and random manner, which may lose essential information. To alleviate this issue, we propose an activation modulation and recalibration scheme to generate high-quality CAMs.

\subsection{Attention Mechanism}
The attention mechanism~\cite{wu2019pseudo,wu2018image} has been widely used in segmentation networks to build the global context relation of images. Non-local~\cite{wang2018non} is the first to take account of the correlation between each spatial point in the feature maps. Then, asymmet~\cite{zhu2019asymmetric} proposes an asymmetric non-local network to strengthen the connection of non-local networks. SE~\cite{hu2018squeeze} learns the importance of channel features by computing the interactions between channels. Following this work, ~\cite{Wang2020ECANetEC} uses a channel-based convolution to learn the interactions. CBAM~\cite{woo2018cbam} exploits the spatial-wise and the channel-wise attention to highlight the important cues in the channel and spatial dimension. ~\cite{cao2019gcnet} incorporates long-range dependencies to the fundamental attention module. In this paper, we introduce an attention modulation module to enhance the minor but essential features for the segmentation task. 

\section{Methodology}
In this section, we first briefly introduce the conventional method for CAMs generation. Then we illustrate the activation modulation and recalibration scheme (AMR). The motivation and details of the proposed AMM is introduced in the next section. Finally, the modulation function and training loss functions are illustrated.

\subsection{Preliminary}
Class Activation Maps (CAMs)~\cite{zhou2016learning} denote the response regions of specific classes for the input images $I \in \mathcal{R}^{3 \times H \times  W}$. A multi-label classification network is employed for encoding the features of all classes, which can be leveraged to extract the feature maps $F(I) \in \mathcal{R}^{C \times H \times  W}$ before the last classification layer to obtain CAMs.  $C$ indicates the channel numbers of features maps. Then we simply perform matrix multiplication on $F(I)$ to generate CAMs:
\begin{equation}
    M(I) = w_N^TF(I) \text{,}
 \end{equation}
where $M(I) \in \mathcal{R}^{N \times H \times  W}$ is the obtained CAMs. $w_N^T$ is the weight of the last fully-connected layer for $N$ classes. 

However, such CAMs are classification-oriented and ignore the task-specific of semantic segmentation.
Namely, the network is optimized via classification-based loss, which resorts to some discriminative regions of the full objects to accomplish the classification task.  It will sacrifice the performance of weakly supervised semantic segmentation, which needs to obtain the holistic bound of the whole object.
To address this issue, we propose the Activation Modulation and Recalibration (AMR) scheme to recalibrate initial CAMs to be more task-specific.

\subsection{Activation Modulation and Recalibration Scheme}

We illustrate the activation modulation and recalibration (AMR) scheme in Fig.~\ref{fig: framework}. The AMR consists of the spotlight branch and the compensation branch. The spotlight branch is similar to the previous methods~\cite{wei2017object,jiang2019integral, lee2021anti}, which employs the classification loss to optimize itself and generate the spotlight CAMs $M_S$. Because the spotlight branch frequently activates the informative features during the training procedure, the obtained CAMs mainly highlight the discriminative regions of target objects.  

The compensation branch is craftily designed to play as auxiliary supervision for the spotlight CAMs. It alleviates the task gap issue of using classification-based CAMs to perform segmentation tasks in previous work, which contributes to providing more semantic segmentation-special cues. The compensation branch can be regarded as a  plug-and-play component, which can dig out the essential regions for segmentation that are easily ignored by the spotlight branch. The obtained compensation CAMs $M_C$ helps to
recalibrate the spotlight CAMs $M_S$ to generate the final weighted CAMs $M_W$, which is illustrated as:
\begin{equation}
    M_W(I) = \xi  M_S(I) + (1-\xi) M_C(I) \text{,}
 \end{equation}
where $\xi$ denotes the recalibration coefficient.

\subsection{Attention Modulation Module}
The attention modulation module (AMM) is proposed to assist the compensation branch to extract more regions essential for semantic segmentation tasks. As shown in Fig~\ref{fig: framework}, AMM consists of channel attention modulation and spatial attention modulation. We firstly feed features $F(I)$ to the channel AMM. The channel interdependencies are explicitly modeled by the average pooling and the convolutional layer, which reflect the sensitivity to informative features. Inspired by~\cite{jiang2019integral}, the most sensitive features correspond to the discriminative regions, the minor features denote the important but easily ignored regions, and the insipid features may indicate the background concepts. Therefore, we exploit a modulation function to enhance the minor features and restrain the most and least sensitive features. The above operations can be denoted as:
\begin{equation}
    A_c = \mathcal{G} (H(P_s(F(I)))) \text{,}
 \end{equation}
where $A_c$ is the channel attention map. We denote $P_s$ as the spatial average pooling function and $H$ as the convolution layer. Then the modulation function $\mathcal{G} $ is leveraged to reassign the distribution of features to highlight the minor features in the the channel dimension. 

Then we conduct an element-wise multiplication between the channel attention maps and input feature maps to generate the redistributed features, which is defined as,
\begin{equation}
    F_c(I) = \tilde{A_c} \odot F(I) \text{,}
 \end{equation}
where $\tilde{A_c}$ denotes the channel attention maps which are expanded to the dimensions of feature maps. $F_c(I)$ represents the output feature maps.

To further model inter-spatial relationship in the spatial dimension, we also introduce a spatial AMM to cascade after the channel AMM. Specifically, we first employ a channel average pooling $P_c$ on $F_c(I)$ in channel dimension and then apply a convolution operation $H$  to them. The output feature maps illustrate the importance of the features in the spatial dimensions. Then we perform a modulation function on the output feature maps to increase the minor activations. The implementation process can formulate as:
\begin{equation}
    A_s = \mathcal{G} (H(P_c(F_c(I))))\text{,}
 \end{equation}
where $A_s$ is the spatial attention map. The high activation values in $A_s$ reflect the easily ignored regions. Then we make an element-wise multiplication between the spatial attention maps and the feature maps:
\begin{equation}
    F_s(I) = \tilde{A_s} \odot F_c(I)\text{,}
 \end{equation}
where $\tilde{A_s}$ denotes the spatial attention maps that are expanded to the dimensions of feature maps. 

\begin{figure}[t]
    \centering
    \begin{center}
        \includegraphics[width=0.99\columnwidth]{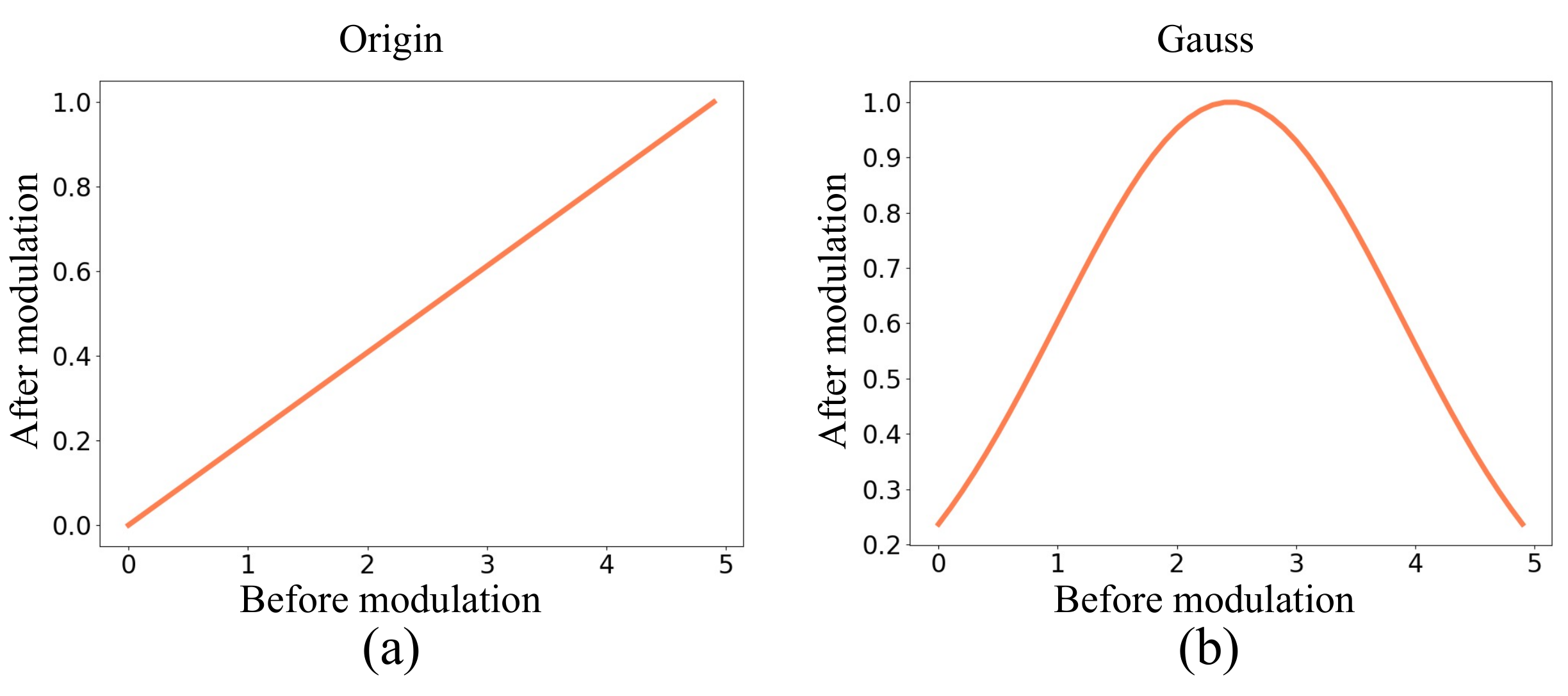}
    \end{center}
    \caption{Illustration of modulation function. The values on the axis denote the range of distribution of activations. (a) represents the original activation distribution. (b) indicates the redistributed activations, which are modulated by the gaussian function to highlight the minor activations.}
    \label{fig: projection}
    \end{figure}

\subsection{Modulation Function}
In AMM, we employ the modulation function to redistribute the activation values of feature maps: 
    \begin{equation}
    \mathcal{V}_A = \mathcal{G} (\mathcal{V}_{A_f}) \text{,}
    \end{equation}
where $\mathcal{G}$ represents the gaussian function, which maps all activation values into a gaussian distribution. The parameters of ``mean'' and ``std'' are calculated by the values of $\mathcal{V}_{A_f}$:
\begin{equation}
    \mu = \frac{1}{M} \sum_{i=1}^{M}(\mathcal{V}_{A_f}^i)  \text{,} \quad \sigma = \sqrt{ \frac{1}{M} \sum_{i=1}^{M}(\mathcal{V}_{A_f}^i - \mu)^2} \text{,}
\end{equation}

where $\mu$ and $\sigma$ are the mean and standard deviation of activation maps. We follow the seting of $\mu$ and $\sigma$ to project the activation values in $\mathcal{G}$.

We visualize the distribution of activations before and after the modulation in Fig.~\ref{fig: projection}. We observe that the gaussian projection greatly suppresses the most and the least important activations. And it emphasizes the minor activations to extract the easily-ignored regions directly, which is crucial for the segmentation task. In addition, we also explore directly set the thresholds to change the importance distribution. But it is difficult to determine an uniform threshold for all images. The experimental results of different modulation functions are summarized in Tab.~\ref{tab: project_comp}.

\subsection{Loss Function}
In the training procedure, we employ a global average pooling operation and a fully-connected layer to obtain the prediction $Y$, which represents the class probability for all categories. Finally, we leverage the multi-label soft margin loss $\mathcal{L}_{cls}$ to optimize $Y$:
\begin{equation}
    \mathcal{L}_{cls} = -\frac{1}{N} \sum_{i=1}^{N}(\tilde{Y_i}log(\frac{1}{1+e^{-Y_i}}) + (1-\tilde{Y_i})log(\frac{e^{-Y_i}}{1+e^{-Y_i}})) \text{,}
 \end{equation}
where $N$ denotes the number of classes and $\tilde{Y}_i$ denote the label of the category $i$.
We provide two classification losses to supervise two classification heads in the AMR. The $\mathcal{L}_{cls}^s$ indicates the supervision of the spotlight branch. And the $\mathcal{L}_{cls}^c$ is supervised for the compensation branch. In short, the total classification loss can be illustrated as:
\begin{equation}
    \mathcal{L}_{cls} = \frac{1}{2}(\mathcal{L}_{cls}^s + \mathcal{L}_{cls}^c) \text{.}
 \end{equation}

 To make full use of complementary CAMs from the counterpart branch, we employ a cross pseudo supervision on the spotlight CAMs and the compensation CAMs. It can be viewed as a semantic similar regularization for each branch:
\begin{equation}
    \mathcal{L}_{cps} = \ \parallel M_S - M_C \parallel _1 \text{,}
 \end{equation}
where  $\mathcal{L}_{cps}$ not only regularizes the compensation branch but also pulls the discriminative regions and easily ignored regions close to each other. Therefore, we can obtain two complementary regions as seeds to recalibrate the initial CAMs.
To sum up, the proposed AMR is optimized with the final loss function $\mathcal{L}_{all}$:
\begin{equation}
    \mathcal{L}_{all} = \mathcal{L}_{cls} + \mathcal{L}_{cps} \text{.}
 \end{equation}
 
\section{Experiment}

\begin{table}[t!]
    \centering
    \begin{threeparttable}
        \resizebox{\linewidth}{!}{
          \begin{tabular}{lccc}
          \toprule
          \textbf{Methods} & \textbf{Sup. } & \textbf{Val} & \textbf{Test}  \\ 
          \midrule
          AffinityNet~\cite{ahn2018learning} & $\mathcal{I}$  & 61.7 & 63.7 \\
          IRNet~\cite{ahn2019weakly} & $\mathcal{I}$ & 63.5 & 64.8 \\
          CIAN~\cite{fan2020cian} & $\mathcal{I}$  & 64.3 & 65.3 \\
          SSDD~\cite{shimoda2019self} & $\mathcal{I}$  & 64.9 & 65.5 \\
          OAA+~\cite{jiang2019integral} & $\mathcal{I}$ & 65.2 & 66.9 \\
          SEAM~\cite{wang2020self} & $\mathcal{I}$  & 64.5 & 65.7 \\
          Chang \emph{et al}.~\cite{chang2020weakly} & $\mathcal{I}$  & 66.1 & 65.9 \\
          Zhang \emph{et al}.~\cite{zhang2020reliability} & $\mathcal{I}$  & 66.3 & 66.5 \\
          Chen \emph{et al}.~\cite{chen2020weakly} & $\mathcal{I}$  & 65.7 & 66.7 \\
          CONTA~\cite{zhang2020causal} & $\mathcal{I}$  & 66.1 & 66.7 \\
          DRS~\cite{kim2021discriminative} & $\mathcal{I}$ & 66.8 & 67.4 \\
        AdvCAM~\cite{lee2021anti} & $\mathcal{I}$  & 68.1 & 68.0 \\
        \midrule
        MCOF~\cite{wang2018weakly} & $\mathcal{I} +\mathcal{S}$  & 60.3 & 61.2 \\
        SeeNet~\cite{hou2018self} & $\mathcal{I} +\mathcal{S}$ & 63.1 & 62.8 \\
        DSRG~\cite{huang2018weakly} & $\mathcal{I} +\mathcal{S}$ & 61.4 & 63.2 \\
        FickleNet~\cite{lee2019ficklenet} & $\mathcal{I} +\mathcal{S}$ & 64.9 & 65.3 \\
        MCIS~\cite{sun2020mining} & $\mathcal{I} +\mathcal{S}$   & 66.2 & 66.9 \\
        ICD~\cite{fan2020learning} & $\mathcal{I} +\mathcal{S}$ & 67.8 & 68.0 \\
        Yao \emph{et al}.~\cite{yao2021non} & $\mathcal{I} +\mathcal{S}$ & 68.3 & 68.5 \\
          \midrule
          AMR (Ours) & $\mathcal{I}$  & \textbf{68.8} & \textbf{69.1} \\
          \bottomrule
          \end{tabular}}
       \end{threeparttable}
       \caption{Comparison with the state-of-the-art methods on PASCAL VOC2012 val and test set. All results are evaluated in mIoU(\%). $\mathcal{I}$ represents the image-level label and $\mathcal{S}$ indicates the saliency label.}
       \label{tab: sota_comp}
    \end{table}

\subsection{Datasets and Evaluation Metric}
We evaluate our approach on the PASCAL VOC2012 dataset~\cite{everingham2015pascal}. It contains 20 foreground objects classes and one background class. Following the common methods~\cite{wei2017object, wang2020self}, we use 10,582 images for training, 1,449 images for validation, and 1,456 ones for testing. During the whole training process, we only adopt the image-level class labels for supervision. Each image may contain multi-class labels. To evaluate the performance of experiments, we calculate the mean intersection over union (mIoU) of all classes.

\subsection{Implementation Details}
We employ ResNet50~\cite{he2016deep} as the backbone of AMR. We train the network for 8 epochs with a batch size of 16. The initial learning rate is set to 0.01 with a momentum of 0.9. We leverage the stochastic gradient descent algorithm for network optimization with a  0.0001 weight decay. We also take some typical data augmentations on the training images such as random scaling and horizontal flipping. Following the works~\cite{ahn2018learning, ahn2019weakly}, we exploit the random walk algorithm on the obtained CAMs to refine the pseudo labels. After obtained the final pseudo labels for segmentation, we train the DeepLab-v2~\cite{chen2017deeplab} with the backbone of ResNet101~\cite{he2016deep}, which is pre-trained on the ImageNet~\cite{russakovsky2015imagenet}. 

\subsection{Comparison with State-of-the-art Methods}

% \begin{table*}[t]
%     \centering
%     \resizebox{\linewidth}{!}{
%         \begin{tabular}{l|ccccccccccccccccccccc|c}
%         \hline
%         Methods & Bkg & Aero & Bike & Bird & Boat & bottle & Bus & Car & Cat & Chair & Cow & Table & Dog & Horse & Motor & Person & Plant & Sheep & Sofa & Train & Tv & mIoU \\
%         \hline
%         MCOF & 87.0 & 78.4 & 29.4 & 68.0 & 44.0 & 67.3 & 80.3 & 74.1 & 82.2 & 21.1 & 70.7 & 28.2 & 73.2 & 71.5 & 67.2 & 53.0 & 47.7 & 74.5 & 32.4 & 71.0 & 45.8 & 60.3 \\
%         FickleNet & 89.5 & 76.6 & 32.6 & 74.6 & 51.5 & 71.1 & 83.4 & 74.4 & 83.6 & 24.1 & 73.4 & 47.4 & 78.2 & 74.0 & 68.8 & 73.2 & 47.8 & 79.9 & 37.0 & 57.3 & 64.6 & 64.9 \\
%         AffinityNet & 88.2 & 68.2 & 30.6 & 81.1 & 49.6 & 61.0 & 77.8 & 66.1 & 75.1 & 29.0 & 66.0 & 40.2 & 80.4 & 62.0 & 70.4 & 73.7 & 42.5 & 70.7 & 42.6 & 68.1 & 51.6 & 61.7 \\
%         SEAM & 88.8 & 68.5 & 33.3 & 85.7 & 40.4 & 67.3 & 78.9 & 76.3 & 81.9 & 29.1 & 75.5 & 48.1 & 79.9 & 73.8 & 71.4 & 75.2 & 48.9 & 79.8 & 40.9 & 58.2 & 53.0 & 64.5 \\
%         Chang \emph{et al}.  & 88.8 & 51.6 & 30.3 & 82.9 & 53.0 & 75.8 & 88.6 & 74.8 & 86.6 & 32.4 & 79.9 & 53.8 & 82.3 & 78.5 & 70.4 & 71.2 & 40.2 & 78.3 & 42.9 & 66.8 & 58.8 & 66.1 \\
%         \hline
%         Ours & & & \\
%         \hline
%         \end{tabular}}
       
%        \caption{Comparison with SOTA on PASCAL VOC2012 val and test dataset.}
%        \label{tab: comp_each}
%     \end{table*}  

\begin{figure*}[t]
    \centering
    \begin{center}
        \includegraphics[width=0.99\textwidth]{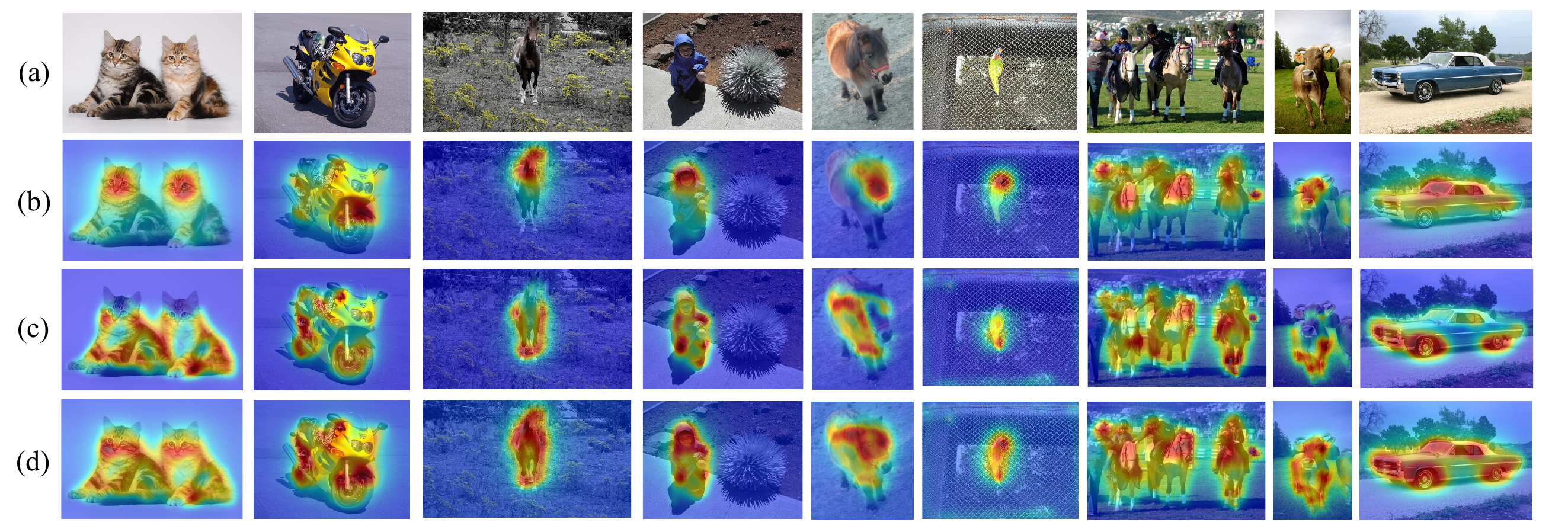}
    \end{center}
    \caption{Visualization of the generated CAMs by our method on the VOC2012 train set. (a) Input images. (b) The spotlight CAMs generated by the spotlight branch. (c) The compensation CAMs generated by the compensation branch. (d) The weighted CAMs incorporated by two complementary CAMs.}
    \label{fig: cams}
    \end{figure*}   

\subsubsection{Comparison on semantic segmenation task.}
We conduct the experiments on the DeepLab v2~\cite{chen2017deeplab} with the obtained pseudo labels of the training set. We report the results on the PASCAL VOC2012 validation and test set, which are shown in Tab.~\ref{tab: sota_comp}. 
On the one hand, AMR significantly outperforms the image-level weakly supervised method and establishes a new state-of-the-art performance. AMR achieves \textbf{68.8\%} of mIoU on the validation set and \textbf{69.1\%} on the test set, which outperforms DRS~\cite{kim2021discriminative} with 2.0\% and 1.7\% respectively. 
On the other hand, AMR even achieves better or comparable results than some algorithms with more granular supervision cues. 
For instance, AMR surpasses the ~\cite{yao2021non} with 0.5\% on validation and 0.6\% on the test set,  which uses the extra saliency supervision. This is an inspiring result as it reveals that our method can get impressive results via learning from massive and cheap annotations, which is of great benefit to practical application.  

\begin{table}[t!]
    \centering
    \begin{threeparttable}
        \begin{tabular}{lcc}
        \toprule
        \textbf{Methods} & \textbf{CAM } & \textbf{Pseudo}  \\ 
        \midrule
        AffinityNet~\cite{ahn2018learning} & 48.0 & 59.7 \\
        IRNet~\cite{ahn2019weakly} & 48.3 & 66.5 \\
        CONTA~\cite{zhang2020causal}  & 48.8 & 67.9 \\
        SEAM~\cite{wang2020self}  & 55.4 & 63.6 \\
        Chang \emph{et al}.~\cite{chang2020weakly}  & 50.9 & 63.4 \\
        \midrule
        AMR (Ours) & \textbf{56.8} & \textbf{69.7} \\
        
        \bottomrule
        \end{tabular}
        \end{threeparttable}
        \caption{Quality results (mIoU) of pseudo labels on the VOC2012 train images. The ``CAM'' column indicates the initial CAM seeds generated by the classification network. The ``Pseudo'' represents the refined pseudo labels used to supervise segmentation.}
        \label{tab: quality_comp}
    \end{table}

\begin{table}[t!]
    \centering
    \begin{threeparttable}
        \begin{tabular}{cccccc}
        \toprule
        \textbf{Baseline} & \textbf{AMM$_c$} & \textbf{AMM$_s$} & \textbf{ $\mathcal{L}_{cps}$ } & \textbf{mIoU(\%)} \\ 
        \midrule
        \cmark &  & & & 48.3 \\
        \cmark & \cmark & & & 52.9 \\
        \cmark &  & \cmark & & 53.5 \\
        \cmark & \cmark & \cmark & & 54.9 \\
        \cmark & \cmark & \cmark & \cmark & \textbf{56.8} \\
        \bottomrule
        \end{tabular}
        \end{threeparttable}
        \caption{Comparison with different effects of each component of our method. The ``Baseline'' represents a single classification network. The ``AMM$_c$'' and ``AMM$_c$'' denote the proposed channel AMM and spatial AMM respectively. $\mathcal{L}_{cps}$ denotes the semantic regularization.}
        \label{tab: ablation_comp}
    \end{table}

\subsubsection{Comparison on CAM and pseudo labels.}
The proposed scheme aims to provide segmentation-specific CAMs to improve the quality of the pseudo labels. In order to verify the effectiveness of our method in generating CAMs and pseudo labels, we summarize the results of the CAMs and the pseudo-labels of the PASCAL VOC2012 training set with several competitive methods (see Tab.~\ref{tab: quality_comp}). It reveals that the AMR achieves the mIoU of \textbf{56.8\%} and \textbf{69.7\%} in terms of CAM and pseudo labels, respectively. Our method surpasses the advanced method SEAM~\cite{wang2020self} with 1.4\% in CAM and outperforms the CONTA~\cite{zhang2020causal} by 1.8\% in pseudo labels. 
Note that SEAM~\cite{wang2020self} uses Wide ResNet38~\cite{wu2019wider} as the backbone, which achieves superior performance than ResNet50 in their work. The experimental results indicate that our compensation CAMs can effectively improve the quality of the initial CAMs and pseudo labels. 

To illustrate how AMR improves the quality of pseudo labels, we visualize the CAMs provided by AMR in Fig.~\ref{fig: cams}. 
From this figure, we have the following observations.
i) The spotlight CAMs generated by the spotlight branch mostly focuses on the discriminative regions. 
ii) The compensation CAMs highlight the regions that are essential for targets but easily ignored. It dues to the fact that AMM helps to modulates the activation maps to emphasize the minor features. 
iii) The weighted CAMs contain more complete regions than spotlight CAMs, which is consistent with the essence of the semantic segmentation task.

\subsection{Ablation Studies}

\subsubsection{Effectiveness of core components.}
To verify the effectiveness of core components in our approach, we increase each essential component gradually on the basis of the single classification network (abbreviated as ``baseline") that only contains the spotlight branch.
We compare the performance of different components with the variant ``baseline" in Tab.~\ref{tab: ablation_comp}.
 As shown in Tab.~\ref{tab: ablation_comp}, AMM$_c$ and AMM$_s$ improve the mIoU of CAMs to 52.9\% and 53.5\% respectively. And the whole AMM achieves 54.9\%.
Furthermore, the cross pseudo supervision $\mathcal{L}_{cps}$ contributes to achieving 1.9\% performance improvement. The whole framework achieves the best performance 56.8\%. These ablation experiments demonstrate the effectiveness of each core component in our method.

\begin{table}[t!]
    \centering
    \begin{threeparttable}
            \begin{tabular}{lccc}
            \toprule
            \textbf{Methods} &   \textbf{Baseline} &    \textbf{Threshold} &  \textbf{Gauss} \\ 
            \midrule
            \textbf{mIoU(\%)} & 48.3 & 50.1 & \textbf{56.8} \\
            
            \bottomrule
            \end{tabular}
        \end{threeparttable}
        \caption{Comparison with different modulation functions. mIoU is evaluated on the CAMs of VOC2012 train images.}
        \label{tab: project_comp}
    \end{table}     

    \begin{figure*}[t!]
        \centering
        \begin{center}
            \includegraphics[width=0.99\textwidth]{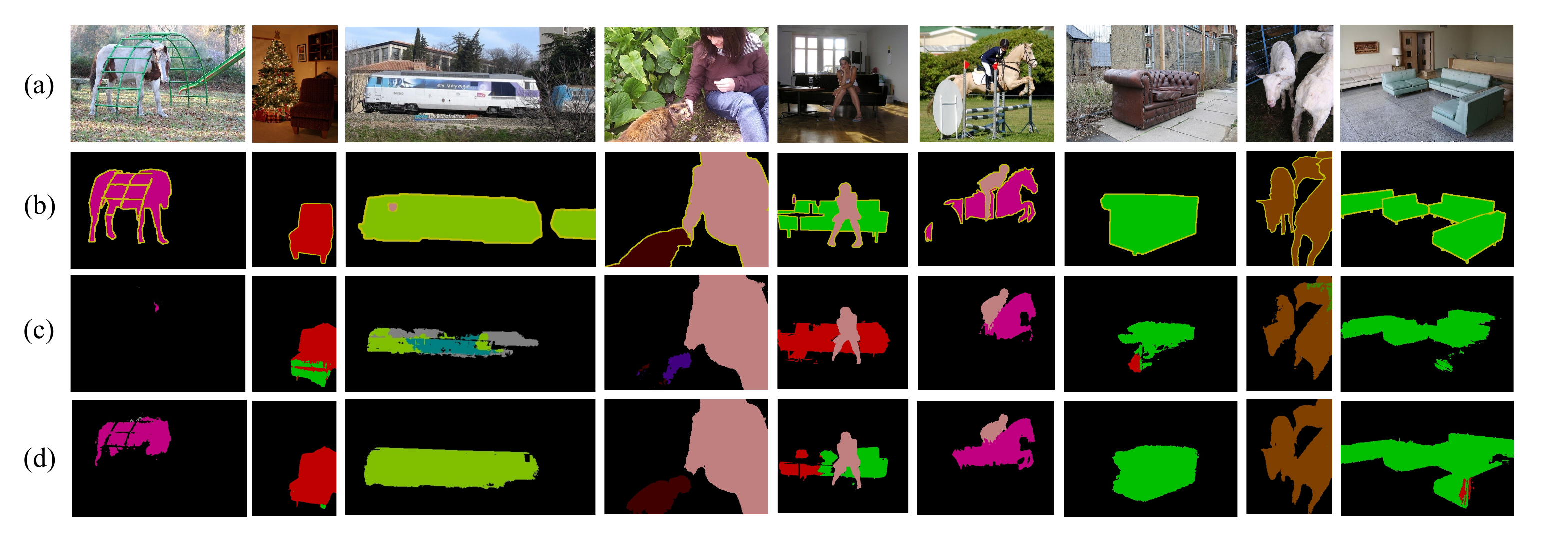}
        \end{center}
        \caption{Qualitative results on the PASCAL VOC2012 validation set. (a) Input images. (b) Ground truth labels. (c) The segmentation results by IRNet~\cite{ahn2019weakly}. (d) The segmentation results of our approach.}
        \label{fig: val_seg}
        \end{figure*}

\subsubsection{Effectiveness of modulation functions.}
In Tab.~\ref{tab: project_comp}, we compare the results of different modulation functions introduced in Fig.~\ref{fig: projection}. 
``Threshold'' modulates the activation to 1 when exceeding the threshold and sets to 0 when the activation is lower than the threshold, which can obtain 1.8\% improvement on the baseline as it remains the most important feature and strengthens some minor activations. The ``Gauss'' function achieves 56.8\% mIoU, which ranks first in all candidate functions. It may because the gaussian function can redistribute the activation maps appropriately to mine some essential concepts that are easy to be ignored.

\begin{table}[t!]
    \centering
    \begin{threeparttable}
            \begin{tabular}{cccccc}
            \toprule
            \textbf{$\xi$} &   \textbf{0.1} &  \textbf{0.3} &  \textbf{0.5} &  \textbf{0.7} & \textbf{0.9}\\ 
            \midrule
            \textbf{mIoU (\%)} & 49.2 & 53.4 & \textbf{56.8} & 54.5 & 50.7\\
            
            \bottomrule
            \end{tabular}
        \end{threeparttable}
        \caption{Comparison with different recalibration coefficient. mIoU is evaluated on the CAMs of VOC2012 train images.}
        \label{tab: weight_comp}
    \end{table} 

\subsubsection{Effectiveness of recalibration coefficient.}
To explore the optimal recalibration coefficient ($\xi$), we report the results in Tab.~\ref{tab: weight_comp}. $\xi$ indicates the contribution of spotlight CAMs to the weighted CAMs. We observe that when setting $\xi$ as 0.5 can achieve the best result, \emph{i.e.} 56.8\%. When increasing or decreasing the value of $\xi$, the performance decreases dramatically, it may be due to the fact that it breaks the balance of regions compensation of two CAMs. When the coefficient is close to 0.1 or 0.9, the framework approximates a single branch, which brings dramatical performance degradation.

\subsection{Generalization Discussion}
To verify the generalization of AMR, we extend the proposed AMR into two advanced methods, \emph{i.e.} IRNet~\cite{ahn2019weakly} and SEAM~\cite{wang2020self}. We remain the original training settings in their paper and compare the results of the initial CAMs.
As shown in Tab.~\ref{tab: transfer_comp}, our approach achieves 8.5\% mIoU improvement on IRNet. 
For that baseline SEAM, we transform the classification backbone to Wide ResNet38~\cite{wu2019wider} as the same with SEAM. The results indicate that AMR improves the quality of CAMs by 2.5\%, which demonstrates the generalization and robustness of our method for incorporating with other approaches to improve segmentation-based CAMs. 

\begin{table}[t!]
    \centering
    \begin{threeparttable}
            \begin{tabular}{lc}
            \toprule
            \textbf{Methods} & \textbf{CAM(mIoU) } \\ 
            
            \midrule
            IRNet~\cite{ahn2019weakly} & 48.3 \\
            IRNet+Ours &  \textbf{56.8} \\
            \midrule
            SEAM~\cite{wang2020self} &  55.4 \\
            SEAM+Ours &  \textbf{57.9} \\
            \bottomrule
            \end{tabular}
        \end{threeparttable}
        \caption{Generalization results of AMR on IRNet~\cite{ahn2019weakly} and SEAM~\cite{wang2020self}.}
        \label{tab: transfer_comp}
    \end{table}

\subsection{Visualization of Segmentation Results}
As illustrated in Fig.~\ref{fig: val_seg}, we compare our method with IRNet~\cite{ahn2019weakly} on the segmentation results in the validation set of PASCAL VOC2012~\cite{everingham2015pascal}. 
As we can see,  the results of IRNet~\cite{ahn2019weakly}  often fall into misjudgment in some ambiguous regions. On the contrary, our approach success to dig out more regions belonging to the target objects to achieve superior segmentation performance.

\section{Conclusion}
In this paper, we propose a novel activation modulation and recalibration (AMR) scheme for WSSS, which leverages a spotlight branch and a plug-and-play compensation branch to obtain weighted CAMs and provide more semantic segmentation-oriented concepts. 
An AMM module is designed to rearrange the distribution of feature importance from the channel-spatial sequential perspective, which contributes to highlighting some essential regions for segmentation tasks but are easy to be ignored.
Extensive experiments on PASCAL VOC2012 dataset demonstrate that AMR achieves the new state-of-the-art performance of weakly supervised semantic segmentation.

\appendix

\bibliography{AMR}

\end{document}